\documentclass[conference]{IEEEtran}
\IEEEoverridecommandlockouts
\usepackage{graphicx}

\begin{document}
%

\title{WIP: A Unit Testing Framework for Self-Guided Personalized Online Robotics Learning
}
\author{\IEEEauthorblockN{Ponkoj Chandra Shill \hspace{3em} David Feil-Seifer}
\thanks{This material is based upon work supported by the National Science Foundation under Grants IIS-2150394, DUE-2142360.  Any opinions, findings, and conclusions or recommendations expressed in this material are  those of the author(s) and do not necessarily reflect the views of the  National Science Foundation.}
\IEEEauthorblockA{Computer Science and Engineering \\ University of Nevada, Reno,\\ Reno, NV, USA \\ pshill@unr.edu | dave@cse.unr.edu}
\and
\IEEEauthorblockN{Jiullian-Lee Vargas Ruiz}
\IEEEauthorblockA{Computer Science \\ University of Puerto Rico, Arecibo,\\ Arecibo, PR, USA \\ jiullianlee.vargas@upr.edu}
\and
\IEEEauthorblockN{Rui Wu}
\IEEEauthorblockA{Computer Science \\ East Carolina University,\\ Greenville, NC \\ wur18@ecu.edu}
}


%

\maketitle

\begin{abstract}

Our ongoing development and deployment of an online robotics education platform highlighted a gap in providing an interactive, feedback-rich learning environment essential for mastering programming concepts in robotics, which they were not getting with the traditional code-simulate-turn in workflow. Since teaching resources are limited, students would benefit from feedback in real-time to find and fix their mistakes in the programming assignments. To address these concerns, this paper will focus on creating a system for unit testing while integrating it into the course workflow. We facilitate this real-time feedback by including unit testing in the design of programming assignments so students can understand and fix their errors on their own and without the prior help of instructors/TAs serving as a bottleneck. In line with the framework's personalized student-centered approach, this method makes it easier for students to revise, and debug their programming work, encouraging hands-on learning. The course workflow updated to include unit tests will strengthen the learning environment and make it more interactive so that students can learn how to program robots in a self-guided fashion.

\end{abstract}



%
\IEEEpeerreviewmaketitle

\section{Introduction}

\begin{figure}[t]
    \centering
    \includegraphics[width=0.75\linewidth]{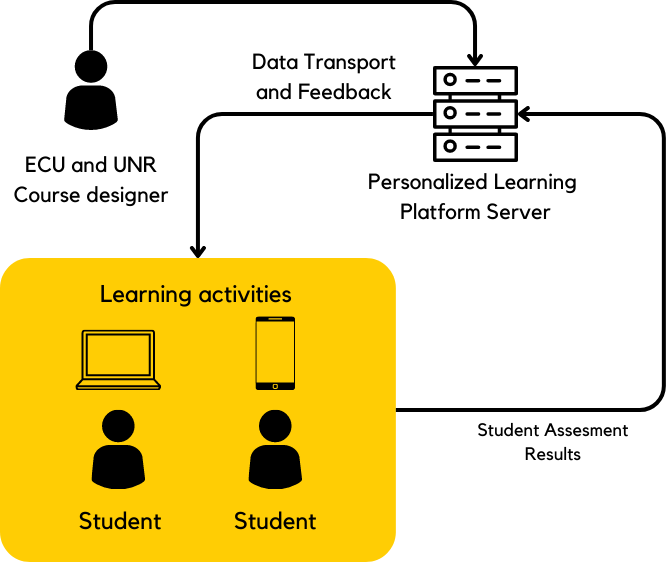}
    \caption{Online Learning Framework: Students can access the learning platform from anywhere. Content and assignments are hosted in a platform server, where the unit testing framework is being Implemented and students learn and do their programming assignments in the platform server as a web-based application and get feedback, grades, and progress through the platform.}
    \label{fig:framework}
\end{figure}

This paper shows work in progress on a unit-testing framework for programming assignments in an online robotics education system. This project aims to increase accessibility to a complete robotics education in remote community institutions and help boost students' performance by extending course content, incorporating mathematical assessments and programming assignments, and, most importantly, implementing a unit testing framework (Fig.~\ref{fig:framework}).
In software development, unit testing refers to serve the purpose of isolating and testing individual components like functions, methods, or objects to verify their performance. Through writing and executing dedicated test cases for each component, developers can detect and rectify errors at an early stage, thereby enhancing code quality and reliability~\cite{sommerville2015software}. 

We are elaborating on self-paced online course materials that can be deployed at community colleges and universities with no robotics-trained experts. This online content will be offered through a remote learning server where students can access it through any device. The students can learn their choice of robotics concepts using an online code editor in a virtual laboratory environment.  A major challenge for robotics courses in general (and for the online robotics system we are developing, in particular~\cite{shill2023wip}) is the labor- and skill-intensive grading process. The proposed unit testing framework will address grading and feedback challenges and assess this framework in our online robotics course environment.

Student feedback from a prior study~\cite{shill2023wip} with our online robotics education system indicated that students were dissatisfied with the inability to properly test their code while doing the assignments or before submission, which led to increased errors and heightened frustration. Moreover, the onerous task of manually grading these programming assessments has proven cumbersome for instructors, creating bottlenecks that delay feedback and undermine the learning process. Unit tests will allow students to receive intermediate feedback while working on programming assignments, help them center their efforts on the graded aspects of the assignments, and give them the ability to test their code in real-time, which might significantly reduce errors. This approach could also help automate the grading process, making it easier for instructors and teaching staff who are less experienced in robotics to provide timely but technically relevant feedback.

In this paper, we show the unit testing framework that will be integrated with the programming assessments activities. These programming activities will be given to the students after completing each module or topic to evaluate their technical progress on robotics learning. We present two study designs to examine this framework in student learning contexts.

\section{background}
Robotics in education provides potential education benefits, positively impacting students by enhancing critical thinking and problem-solving skills \cite{atmatzidou2016advancing, ericson2012effective, greenberg2019exercises, fasola2008robotics}. Delivering quality robotics education is challenging, particularly in community colleges, where a scarcity of qualified robotics instructors exists. Insufficient funding for specialized hardware and software exacerbates this issue~\cite{anderson2019making}. Additionally, students interested in robotics may face financial barriers that prevent them from pursuing this field, given the substantial costs associated with renting or purchasing robots and training programs. We are working to bridge that gap through the creation of a self-directed online robotics learning environment.

A current trend in e-learning is self-directed learning (SDL), where learners take control over the conceptualization, design, implementation, and evaluation of their learning. This corresponds with self-determination theory \cite{deci2012self}, which emphasizes the importance of autonomy, competence, and relatedness for continued intention, and has been proven to be productive in learning environments. Self-direction is dependent on the learners' motivation, self-monitoring, and self-control \cite{sumuer2018factors}. Students have expressed that the process of becoming a self-directed learner is difficult, frustrating and confusing. Therefore, this developmental process should be understood and nurtured; students need a guide for their learning journey that assists them through the process, hence improving their experience and performance \cite{lunyk2001self}. A self-regulated e-learning framework can help reduce participation obstacles and improve students' access to quality robotics education. 

In our earlier work~\cite{shill2023wip}, we introduced a novel strategy to enhance online robotics education by implementing personalized learning approaches.  A personalized mini-course was designed that adapts content and pace to each student's individual needs. The course highlighted hands-on, problem-based learning to cultivate practical skills, including programming assessments, mathematical problem-solving, etc. Moreover, a thorough study was conducted to evaluate the effectiveness of the framework in improving student engagement and learning outcomes. The findings shed light on the scalability and broader application potential of the framework in STEM education \cite{shill2023wip}. However, students also reported being unsatisfied with the educational tools due to technical issues and a lack of means to check or test their work during programming assignments, hence the addition of a TDD-framework for programming assignments.

Test-Driven Development (TDD) is a software engineering practice that revolutionizes traditional programming methodologies by enforcing a `test-first' approach, a concept extensively advocated by Robert C. Martin \cite{martin2011cleancoder}. In TDD, developers begin by writing automated tests for a new feature before any functional code is created. This involves a cycle of writing a test, verifying that it fails, writing the necessary code to pass the test, and refining the code to enhance its structure and efficiency. 
Unit testing is a crucial step that involves examining individual components, such as methods or object classes, to guarantee their proper functioning before being incorporated into larger systems~\cite{sommerville2015software}, identifying errors early in the development process, which simplifies debugging and enhances the overall quality of the software. Unit testing not only promotes rigorous validation of software behavior under predictable conditions but also under simulated events, utilizing mock objects when necessary. Automating unit testing through frameworks is particularly effective, as it enables rapid testing and feedback cycles, thus reducing the risks associated with integration and helping to maintain high standards of software reliability and maintainability.

\begin{figure}[t]
    \centering
    \includegraphics[width=0.90\linewidth]{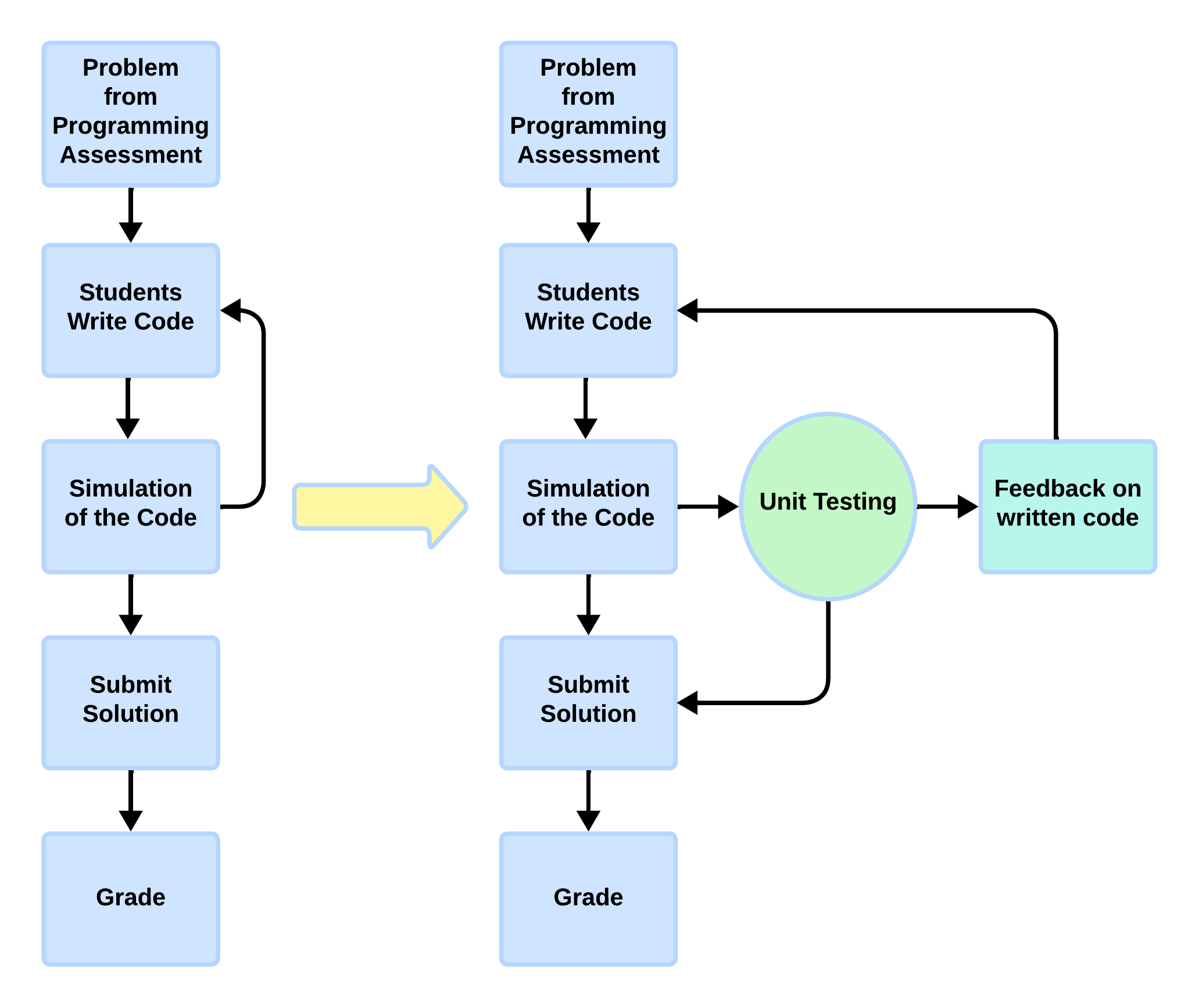}
    \caption{Left- The Prior System Workflow; Right- Proposed Workflow incorporating Unit Testing Framework, in this system, students can test their code and get intermediate feedback, which helps them iteratively improve their robot's behavior and their programming skills prior to submitting the assignment}
    \label{fig:system}
\end{figure}

\section{Approach}

We designed a test-driven development (TDD) framework to provide students with predefined unit tests that accompany each programming task for every topic. These tests will be embedded within our web-based virtual laboratory~\cite{wu2024ASEE}, integrating into the overall workflow. In this section, we present the current course module design, as well as our TDD-based changes to this course workflow, including the feedback that we can provide to students, shown in Fig.~\ref{fig:system}.

\subsection{Virtual Laboratory}

\begin{figure}[t]
    \centering
    \includegraphics[width=0.95\linewidth]{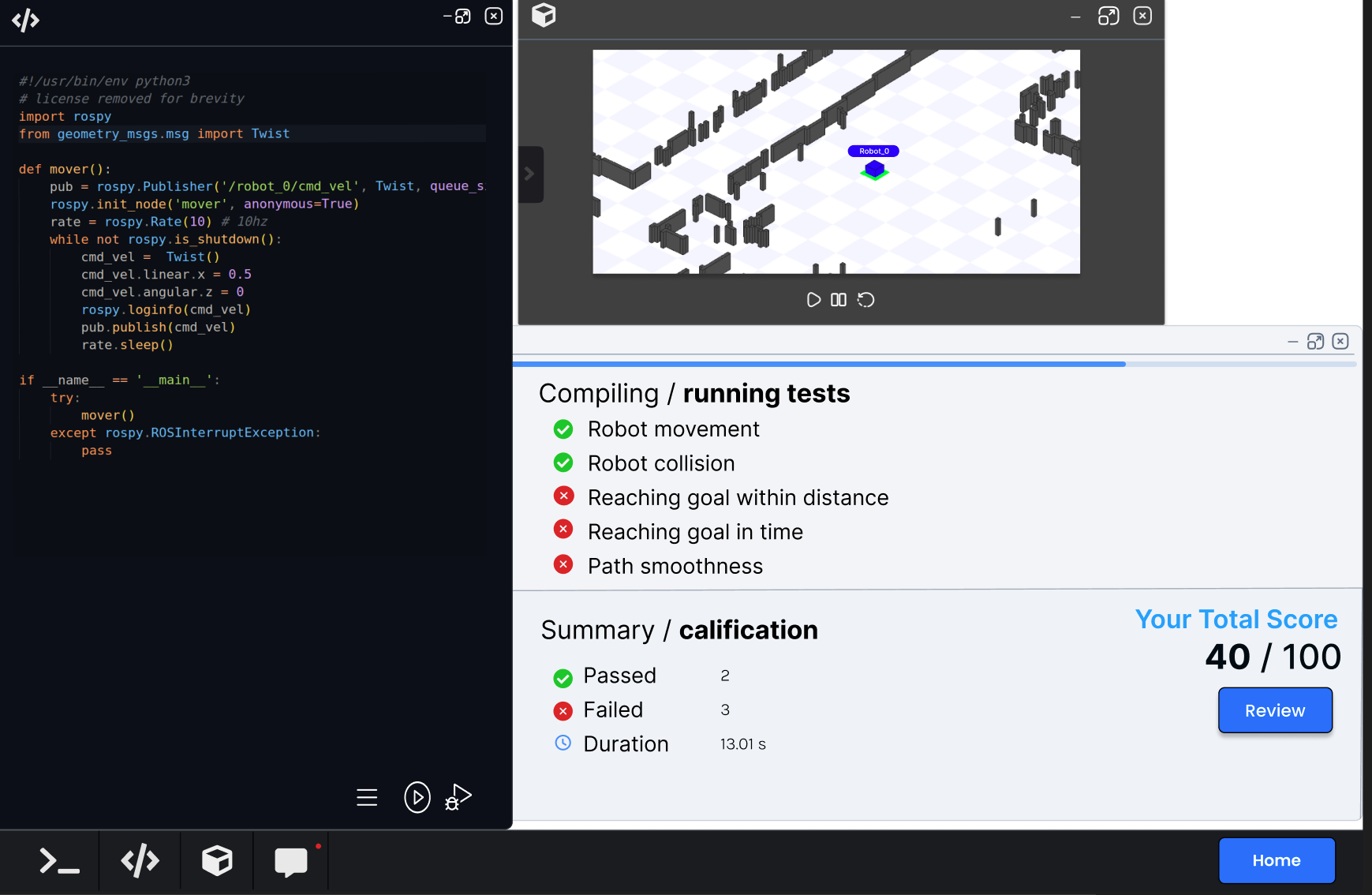}
    \caption{Prototype of the virtual laboratory interface. The virtual laboratory includes a code editor on the left, real-time simulation on the top right, and a feedback window on the bottom right. Students get intermediate feedback, grades, and real-time simulations in the same terminal.}
    \label{virtual_lab}
\end{figure}

This module is where students will complete their programming assessments. After finishing each course topic, students will be redirected to the web-based application, where they will encounter programming assignments presented as questions or problems. The platform will feature a split-screen interface, with the code editor on the left side and a virtual robot or object on the right, visually simulating or demonstrating the code's effect in real-time, shown in Fig.~\ref{virtual_lab}. 

In this virtual environment simulates real-world robots and their sensors in a web-based environment. For example, in the programming assignment for the obstacle avoidance lesson, students can program the robot to navigate through complex environments, avoiding obstacles and following specified paths. This includes activities such as maze solving, where the robot must find its way through a maze using various algorithms, and terrain traversal, where the robot navigates different types of surfaces and elevations. The virtual robot will also be capable of performing sensor-based tasks, allowing students to integrate and process data from virtual sensors like ultrasonic, infrared, or LIDAR sensors. 

For the robot to execute path planning and motion control activities, students would have to program it to calculate and follow the most efficient route from one point to another, considering various constraints and dynamic changes in the environment. This will help students learn about path-finding and motion control algorithms, such as A* or Dijkstra's algorithms or bug algorithms. The virtual laboratory also supports interactive debugging and iterative development. As students write their code, they will see the robot's actions in real time on the simulation screen. If the robot's behavior deviates from expectations, students can immediately adjust their code and observe the changes, fostering a deeper understanding of the relationship between code and robotic behavior. 


\subsection{Assessment}
Programming assignment assessments cover theoretical, mathematical, and practical programming tasks to comprehensively assess students' understanding. For example, in the ``Obstacle Avoidance" topic, students encounter various programming assignments. One example is a scenario where they must program a mobile robot using the Bug algorithm~\cite{xu2017mobile} to navigate a 2D environment. The starting and goal points and the obstacle's position are provided. The robot must follow the obstacle's edge until it can resume its path to the goal. It always turns right when encountering the obstacle and follows the Bug algorithm perfectly.

Students need to calculate the total path length the robot takes to reach the goal, including the detour. They must provide their answer as an expression that includes the perimeter of the obstacle the robot follows and the straight-line paths before and after the detour.
The major items being assessed in this assignment include the student's ability to:

\begin{itemize}
    \item Implement the bug algorithm correctly to navigate around obstacles.
    \item Ensure writing code that the robot consistently follows the edge of the obstacle without collisions.
    \item Calculate the total path length accurately, including both the detour around the obstacle and the straight-line paths.
    \item Apply mathematical concepts to provide a precise expression of the path length.
    \item Debug and refine their code based on the real-time feedback provided by the unit testing framework.

\end{itemize}


\subsection{Unit testing framework}

We modify the above workflow to include a unit testing framework to evaluate the provided code or programming assignments. We integrate a series of unit tests conducted at intermediate points and upon submission of the assignment to evaluate the program's functionality. This framework will be able to provide immediate feedback, highlighting errors and offering suggestions as students write their code. 
For example, in the obstacle avoidance assignment using bug algorithm, the unit testing is designed in such a way that it checks different aspects of the solution to the problem:




\begin{itemize}
  \item The robot does not hit any obstacles while it moves;
  \item The robot does not stop at all;
  \item The robot makes right turns only at the edge of obstacles;
  \item The robot is moving smoothly; and
  \item The robot has reached the goal point.
\end{itemize}

If a coding error induces unwanted behavior, such as the robot hitting an obstacle or stopping prematurely, the unit test will identify these issues. For instance, if the robot's code causes it to collide with an obstacle, the unit test designed to verify obstacle avoidance will fail. This failure will be highlighted to the student in the simulation panel and code editor both, indicating that the robot's movement logic needs revision. The test results will show which specific conditions were not met, providing clear indicators of where the code deviates from expected behavior. This immediate notification of test failures helps students pinpoint and correct their mistakes, facilitating a deeper understanding of programming concepts and improving their coding skills.

\subsection{Feedback System}

\begin{figure}[t]
    \centering
    \includegraphics[width=0.95\linewidth]{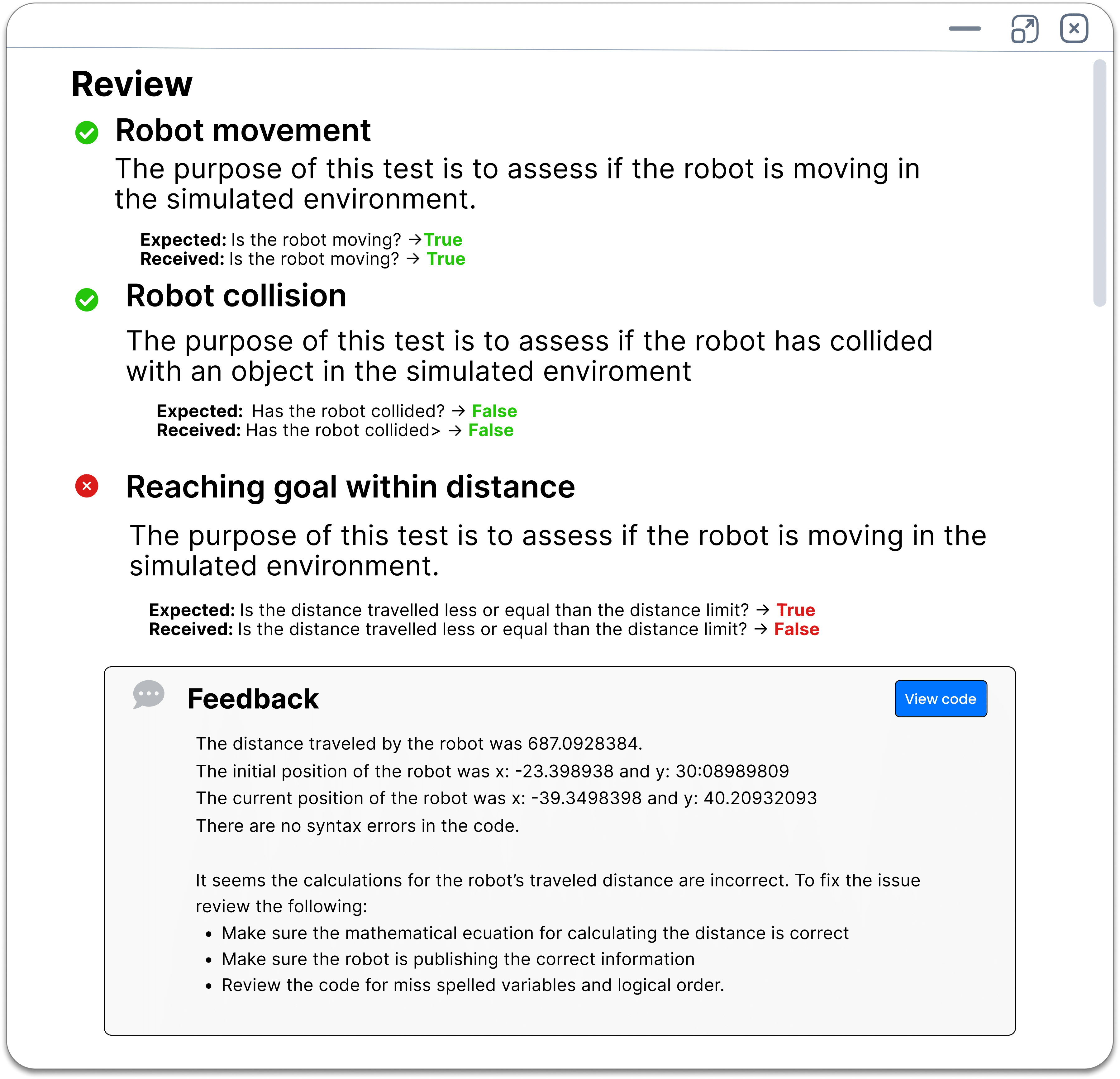}
    \caption{Feedback window prototype showing test results. Students can review the details of the executed tests. The system will show the purpose, requirements and outputs of the tests. On the other hand, the system provides feedback on how the student can verify and correct potential errors. }
    \label{feedback}
\end{figure}

The unit testing system incorporates with feedback system, which provides immediate, objective feedback on the assignment goals. This feedback will be presented in a window that displays the executed tests and indicates which ones passed or failed, shown in Fig.~\ref{feedback}. Students will also have the option to review the details of each test for further understanding and improvement. Moreover, the detailed feedback from each test allows students to understand specific areas of strength and weakness, fostering a deeper comprehension of programming concepts and improving their ability to write robust code.

\section{Proposed Validation and Study}

We first plan a formative study to demonstrate the proper operation of the TDD-based feedback system and its overall utility in the virtual laboratory environment. We plan to do a between-participants study. The control group will use the traditional virtual lab environment to complete a programming assignment; the experimental group will use the test-driven virtual lab environment. We will compare the resultant programs for each group, looking at:

\begin{itemize}
    \item program correctness;
    \item number of iterations testing code before submission;
    \item student sentiment with the programming experience; and
    \item assignment grader ease with grading.
\end{itemize}

These above items will help assess how well the students perform, how long it takes them to complete assignments, how students feel about the experience, and whether graders feel the test-based method is more efficient.


A comprehensive validation of the proposed system, featuring the virtual laboratory and the integrated unit testing framework, will be conducted with participants enrolled in a full-semester self-paced robotics course. Prior to the course, students will undergo a diagnostic test to assess their baseline knowledge of mathematical concepts and robotics. Additionally, a survey will be administered to gauge students' prior experiences and preferences in online courses.



To assess the effectiveness of the proposed system, a regular class will serve as a control group to compare the results of students using the proposed system to those in a traditional linear robotics course. Before the start of the full course, both groups will take the same diagnostic test and pre-survey. After the course ends, they will take the same summative assessment and post-survey. The analysis will assess student satisfaction, subject matter mastery, and the level of autonomy in learning to gauge the effectiveness of the proposed approach.

A post-course survey will also include a section to gather feedback on the effectiveness of the unit testing framework in assisting students with their assignments. The survey will assess students' motivation and satisfaction with the test-driven development (TDD) approach and the virtual lab web-based application implemented in the self-paced course.

This validation study aims to provide valuable insights into the efficacy of the self-paced robotics course compared to the traditional linear approach, based on concrete data and participant feedback. By synchronizing the evaluation of the TDD approach, we will be able to determine how the continuous, real-time feedback and iterative problem-solving facilitated by the unit testing framework impact student learning outcomes and overall course satisfaction.

\section{Conclusion and Future Direction}

This paper demonstrates the development of a unit testing framework within a self-guided, personalized online robotics learning environment. The integration with the virtual laboratory  addresses significant challenges in online robotics education, such as providing timely and actionable feedback, enhancing the learning experience, and reducing the manual grading burden on instructors. By offering real-time, interactive feedback through the unit testing framework, students can iteratively debug and refine their code, leading to a deeper understanding of robotics programming concepts and improved problem-solving skills.

The proposed system's effectiveness will be validated through a comprehensive study comparing the self-paced robotics course with a traditional linear robotics course. Key factors such as student satisfaction, mastery of the subject matter, and level of autonomy in learning will be assessed to gauge the impact of the unit testing framework and the virtual lab on student learning outcomes. This study aims to provide valuable insights into the efficacy of the proposed approach, offering concrete data and participant feedback to support the broader application of such systems in robotics education.

\clearpage

\bibliographystyle{IEEEbib}
\bibliography{refs}

\end{document}